# Dense 3D Regression for Hand Pose Estimation

Chengde Wan[1], Thomas Probst[1], Luc Van Gool[1,3], and Angela Yao[2]

[1]ETH Zürich  [2]University of Bonn  [3]KU Leuven

## Abstract

*We present a simple and effective method for 3D hand pose estimation from a single depth frame. As opposed to previous state-of-the-art methods based on holistic 3D regression, our method works on dense pixel-wise estimation. This is achieved by careful design choices in pose parameterization, which leverages both 2D and 3D properties of depth map. Specifically, we decompose the pose parameters into a set of per-pixel estimations, i.e., 2D heat maps, 3D heat maps and unit 3D directional vector fields. The 2D/3D joint heat maps and 3D joint offsets are estimated via multi-task network cascades, which is trained end-to-end. The pixel-wise estimations can be directly translated into a vote casting scheme. A variant of mean shift is then used to aggregate local votes while enforcing consensus between the the estimated 3D pose and the pixel-wise 2D and 3D estimations by design. Our method is efficient and highly accurate. On MSRA and NYU hand dataset, our method outperforms all previous state-of-the-art approaches by a large margin. On the ICVL hand dataset, our method achieves similar accuracy compared to the nearly saturated result obtained by [5] and outperforms various other proposed methods. Code is available online[1].*

## 1. Introduction

Vision-based hand pose estimation has made significant progress in recent years. The increased performance can be attributed to two dominating trends: depth imaging and deep learning. First of all, hand pose estimation techniques have shifted almost entirely to using only depth inputs[33, 31, 27, 52] since commodity depth sensors such as the MS Kinect and Intel Realsense have become widely available. As a 2.5D source of information, depth significantly resolves much of the ambiguities present in monocular RGB input. Secondly, deep learning has fundamentally transformed the way that vision problems are being solved.

The use of deep neural networks has become the norm for hand pose estimation[43, 25, 51, 10].

In standard hand pose estimation pipelines, depth maps are almost always treated as images. This is especially true for deep learning-based approaches, which heavily rely on the machinery of (2D) convolutional neural networks (CNNs). One line of work for 3D hand poses estimation is holistic regression, that is aiming to directly map the depth images to 3D pose parameters such as joint angles or 3D coordinates. It bypasses having to solve for intermediate representations such as 2D coordinates and is able to capture global constraints and correlations among different joints. However, regressing from highly disparate domains such as image and pose is a very challenging learning task. Furthermore, holistic regression cannot generalize to combinations of local evidence such as different individual finger poses and suffers from translational variance and sensitivity to hand bounding box locations.

CNNs have been successfully applied to 2D body pose estimation[23, 44, 47]; in particular, fully convolutional networks (FCNs) can perform pixel-wise joint detection very accurately [23, 47]. This is formulated as a pixel-wise classification of each pixel being the location of a joint. As such, a second line of work in pose estimation tries to create analogous networks for detecting joints in 2D. Through pixel-wise classification, joint detection can exploit local patterns more explicitly than holistic regression, helping the network to learn better feature maps. The 2D detections and 3D regression can then be combined with a multi-task setup [32, 30, 17, 42], either by feeding the 2D detection heat map as an input to a 3D regression network, or by sharing the feature maps between detection and holistic regression. However, there is no guarantee that the regressed 3D joints, if they were to be projected back to 2D, will be in consensus with the original 2D detection heat-map. Moreover, by design, the aforementioned drawbacks of holistic regression are still not eliminated with this line of work. Other works in 2D detection apply inverse kinematics and use a model-based optimization; however, the severe self-occlusion of the hand creates ambiguities which are diffi-cult to resolve and as such, suffers from accuracy problems which are otherwise not present in body pose estimation.

---

[1]https://github.com/melonwan/denseReg

Despite all the drawbacks of working in 2D, we do not want to directly solve a discrete volumetric detection problem with a 3D CNN. This becomes very parameter-heavy and as a result, severely limits the working resolution[11, 29, 21]. Moreover, as input depth maps are only 2.5D, 3D CNNs struggle to resolve the ambiguities caused by the self-occlusion common in hand poses.

At the core of the problem is the mismatch between 2.5D depth data and traditional CNNs, be it in 2D or 3D. By treating depth maps as a 2D image, we can leverage the advances of CNNs, but we still under-utilize the information present. Yet we also want to avoid converting depth information to a volumetric representation due to the computational overhead and the associated ambiguities. To that end, we propose a combined pixel-wise detection and dense regression method for hand pose estimation. Our proposed method enjoys the benefits of 2D FCN-based detection such as translational-invariance and generalization to different finger gesture combinations. At the same time, dense regression allows us to make 3D estimates and benefit from the merits of holistic 3D regression, such as accounting for correlations and skeleton constraints, without having to work in the discrete volumetric domain.

We make two careful design choices in parameterization to stabilize our training and improve regression robustness. First, we work with offsets instead of absolute joint positions, *i.e.* we regress each pixel to a 3D offset of each joint. Joint offsets have been used in previous works[33, 37, 40, 46] and offer invariance towards translation. It also allows us to keep the original spatial resolution in spite of pooling operations in the CNN. Secondly, we re-parameterize the 3D offsets as a heat map and directional unit vector, leading naturally to a joint detection and regression problem to solve for the two respectively. This form of parameterization leverages both the 2D and the 3D geometric properties of a 2.5D depth map. For a given depth map, we use a 2D CNN to capture local surface patterns but also treat the depth map as a set of 3D points to arrive at a final pose estimate in 3D.

To do so, we first extend the 2D detection heat map into 3D, *i.e.*, value of the heat map is inversely proportional to the 3D distance of corresponding point on the depth map to a specific joint. In addition, we predict unit vector fields, where each vector field corresponds to the direction from the point on the depth map to a certain finger joint. Finally, we also detect the joints in 2D, in the form of a projected heat map. We aggregate all of the estimates together with the mean shift algorithm into a global estimate with consensus between the 2D and 3D estimates.

The proposed method is highly accurate and out performs all previous state-of-arts on three publicly available datasets, *i.e.*, NYU[43], ICVL[39] and MSRA[37]. We also compare our method against several baselines that combine holistic regression with 2D joint detection. In these experiments we observe that, unlike in the case of full body pose estimation, those combination strategies can hardly improve holistic regression and are less accurate than our proposed method by a large margin. We attribute this to the depth ambiguity caused by self occlusion, towards which our proposed method is much more robust.

Our contribution can be summarized as follows:

- we formulate 3D hand pose estimation as a dense regression through a pose re-parameterization that can leverage both 2D surface geometric and 3D coordinate properties;

- we provide a non-parametric post-processing method aggregating pixel-wise estimates to 3D joint coordinates; this post-processing explicitly handles the holistic estimation and ensures consensus between the 2D and 3D estimates;

- we implement several baselines to investigate fusion strategies for holistic regression and 2D joint detection in a multi-task setup; such an analysis has never carried out before for hand pose estimation and provides valuable insights to the field.

## 2. Related Works

**Coupling 2D joint detection with 3D estimation** 3D pose estimation based on 2D observations has a long history in computer vision. Early works [35, 50, 34] are mainly based on low level visual cues, *e.g.*, silhouette or optical flow, and use generative models to resolve the depth ambiguity. More recent works have shifted towards mid- and high-level features, *e.g.* 2D joint detection heat maps or representations from CNNs, due to the availability of highly accurate 2D joint detectors [23, 47]. One line of work [32, 30, 17, 42] formulates 3D pose estimation as a regression problem and couples 2D joint detection and 3D regression in a multi-task setup. Others [2, 22, 4, 19, 53, 43, 51, 10] treat 3D estimation as an model-based optimization on top of the 2D joint detections.

Our approach is similar to [32, 30, 17, 42] in that both 2D and 3D estimations are performed in a multi-task setup. However, rather than using a holistic 3D regression, we perform pixel-wise 3D estimation. This type of fusion scheme is translation invariant and can better generalize to different combinations of finger gestures. Like many others, we also use a post-processing, but ours is much simpler with negligible effort when compared to the computationally expensive energy minimization of [2, 22, 19, 43, 51, 10], nearest neighbour search [4], to employing an additional neural network [19, 53].

**Pose Parameterization**  Skeleton models don't necessarily need to be parameterized with 3D joint coordinates. Many works have modelled pose parameters in other spaces to better exploit the skeleton structure. For example, [25, 45] learn a latent space to model the correlation among different joints, while [49, 54, 36] parameterize pose hierarchically, *i.e.*, location of child joint is dependent on its parent joint along the skeleton tree, to leverage dependencies in the skeleton. [22] models skeleton as distance matrix among different joints and [28, 3] formulate pose parameters as heat maps together with offset vector fields to handle multiple instances 2D detection. Ours is inspired by [28, 3] whereas we work on 3D estimation.

**Hand Pose Estimation**  We limit our discussion to deep learning-based methods and refer the reader to [38] for a detailed review of other model-based and random forest-based methods. Deep learning-based methods fall into two camps: two-stage approaches[43, 10, 51] with 2D joint detection followed model-based optimization versus single-stage approaches [25, 26, 11, 45, 24, 12, 5, 6, 9] of holistic pose regression. The current best-performing methods [5, 12, 24] are all single stage, most likely due to the effective exploitation of joint correlations. Our method takes the advantages from both camps and well exploits the 2D and 3D properties of depth maps.

**Offset Regression and Hough Voting**  Several previous works have successfully employed offset regression for localization and pose estimation tasks [33, 37, 40, 46, 16]. Due to their local nature, they offer invariance to translation and their compatibility for bottom-up estimation. However, these methods typically rely on hand-crafted features, with the exception of works on 2D localization [20, 48]. In this work, we extend this idea by learning dense 3D offset regression end-to-end.

## 3. Method

We leverage both the 2D and 3D properties of a depth map to formulate hand pose estimation as a pixel-wise regression problem. From a 2D perspective, we treat the depth map as a 2D surface embedded in 3D and use a convolutional neural network(CNN) composed of 2D convolutional layers to capture surface local geometric patterns. From a 3D perspective, the depth map can also be regarded as a set of 3D points. It is for this set of points that we want to estimate offsets to the hand joints. More specifically, we use a CNN to estimate a dense vector field of offsets for each joint of hand. We re-parameterize the joint offset as a 3D heat map and a directional unit vector and solve for the two via detection and regression respectively (Sec 3.1). The resulting network is fully convolutional and compatible with current joint detection network architectures (Sec 3.2). Several networks can be stacked together as intermediate forms of supervision, with all the estimated results being fed into the next stage to boost pose accuracy. We adopt mean shift (Sec 3.3) to aggregate the pixel-wise regression estimates, while enforcing the 2D projections of the final estimated 3D joints to be in consensus with the pixel-wise 2D joint detections.

### 3.1. Pose Parameterization

Instead of directly regressing 3D joint coordinates from the depth map, like most other regression-based methods [25, 26, 11, 45, 24, 12, 5, 6], we want to estimate an offset vector between depth points and hand joints. This makes the estimate translation-invariant and also generalizes better to different combinations of finger poses. However, directly regressing the 3D offset vector field is non-ideal. First of all, the regression for points that are far from a given hand joint will result in offset vectors with large norms that dominate the training loss. Furthermore, far away hand joints are beyond the scope of the receptive field of the convolutional filters anyway. As such, we decompose the 3D offset vector into two components – a 3D heat map $S$, estimated via detection, and a directional unit vector, $V$, via regression, as follows:

$$S_j(p) = \begin{cases} \theta - \|p - p_j\|_2 & \|p - p_j\|_2 \leq \theta, \\ 0 & \text{otherwise}; \end{cases} \quad (1)$$

$$V_j(p) = \begin{cases} \frac{p - p_j}{\|p - p_j\|_2} & \|p - p_j\|_2 \leq \theta, \\ 0 & \text{otherwise}. \end{cases} \quad (2)$$

where $p \in \mathcal{R}^3$ and $p_j \in \mathcal{R}^3$ are the 3D coordinates of a point from the depth map and of joint $j$ respectively. $\theta$ defines the radius of a 3D ball centered at the joint position that establishes a candidate region(see Fig. 1) from which we consider support. The 3D heat map $S_j(p)$ can be regarded a direct extension of the 2D heat map.

In addition, we estimate the joint's 2D projection as a heatmap $R$,

$$R_j(p) = \begin{cases} \tau - \|\Pi(p) - \Pi(p_j)\|_2 & \|\Pi(p) - \Pi(p_j)\|_2 \leq \tau, \\ 0 & \text{otherwise} \end{cases}, \quad (3)$$

where $\Pi(\cdot)$ denotes the 2D perspective projection function and $\tau$ is the radius of the candidate disk. Even though Eq. 1 and 2 are sufficient to recover the 3D joint location, the over-complete estimation with the 2D projection adds robustness to the local estimate. The 2D projection can be combined with the 3D joint estimate with non-parametric methods, which we elaborate in Section 3.3.

### 3.2. Network architecture

The architecture of the detection and regression network is shown in Fig 1. We use the hourglass network[23] as the backbone because it is highly efficient, though any other joint detection network architecture, *e.g.* [47, 28] could potentially be used. The 2D and 3D joint heat maps and the unit vector fields are estimated by network cascades in a learning multi-task manner. Specifically, for $J$ joints, the network first outputs 2D and 3D joint heat maps with two separate sliding pixel-wise fully-connected layers on top of the output feature map of the hourglass module. Since the unit vector field $V_j$ is correlated with the heat map estimates, we concatenate the heat maps together with the hourglass output feature map to determine the unit vector field. To handle the discontinuity of 3D heat map and unit vector field regressions at surface edges, the initial depth map is also provided via concatenation to the input of the 3D heat map network. Similar to [8, 13], we multiply the binarized depth map as a mask with feature map and concatenate it with the initial feature map. This serves as the input for our unit vector field regression component.

Following the paradigm of [23], we stack together several modules with identical architectures to increase the learning power. Estimates from previous modules are used as inputs to the subsequent ones, while intermediate supervision is applied at the end of each module. Specifically, we define a $L_2$ loss over the $J$ joints from $T$ stacks as follows:

$$\mathcal{L} = \sum_{t=1}^{T} \mathcal{L}_R^{(t)} + \mathcal{L}_S^{(t)} + \mathcal{L}_V^{(t)} \quad (4)$$

$$= \sum_{t=1}^{T} \sum_{j=1}^{J} \|R_j^{(t)} - R_j^*\|^2 + \|S_j^{(t)} - S_j^*\|^2 + \|V_j^{(t)} - V_j^*\|^2,$$

where $R_j^*, S_j^*, V_j^*$ represent the respective ground-truth 2D heat maps, 3D heat maps and vector offsets of joint $j$ and $R_j^t, S_j^t, V_j^t$ are corresponding estimates from $t$th stack.

### 3.3. Inference

During inference, we aggregate all of the pixel-wise estimated evidences into holistic 3D joint coordinates with the mean shift algorithm. By design, this process explicitly ensures consensus between the joint detections in 2D and 3D. Since each joint is estimated with the same mean shift process, we omit the joint index $j$ in this section for simplicity. As shown in Alg. 1, the $N$ nearest points to the joint are selected based on the estimated 3D distance. We only select $K$ because points with larger estimated 3D distances tend to amplify the estimation error of offset direction and thus degrade the recovered 3D joint position estimation.

In addition, we provide a more efficient "unweighted" approximation to Algorithm 1 without the 2D projection (step 4) and replace the weights with $(1 + R) \odot S^2$. Table 1 shows that both strategies have nearly identical results. In practice, we choose 5 nearest points as input to mean shift,

---

$^2\odot$ denotes element-wise multiplication

**Algorithm 1** Mean-shift estimation of one joint

*predefined constants:*
- $\theta$       ▷ 3D distance threshold between point from $D$ to joint
- $K$       ▷ number of points selected as input to mean shift
- $\sigma$       ▷ kernel width of mean shift kernel function
- $N$       ▷ number of mean shift iterations

**Input:**
$D \in \mathcal{R}^{h \times w \times 3} \in \mathcal{R}$   ▷ input point cloud coordinates
*outputs from neural network:*
$R \in \mathcal{R}^{h \times w \times 1}$   ▷ 2D heat map, see Eq. 3
$S \in \mathcal{R}^{h \times w \times 1}$   ▷ 3D heat map, see Eq. 1
$V \in \mathcal{R}^{h \times w \times 3}$   ▷ 3D offset unit vector field, see Eq. 2

1: $P = D + \theta(1 - S) \odot V$   ▷ recover the joint coordinate
2: $I = \text{topK}(S) \in \mathcal{N}^{K \times 2}$   ▷ select top K values' indices
3: $\mathcal{P} = P(I) \subset \mathcal{R}^3$   ▷ fetch estimated 3D joint coordinates
4: $\mathcal{P}_{2d} = \{\Pi(p) | \forall p \in \mathcal{P}\} \subset \mathcal{R}^2$   ▷ 2D projection
5: $\mathcal{W} = R(\mathcal{P}_{2d}) \subset \mathcal{R}$   ▷ fetch corresponding 2D heat map values as weights
6: $p = \text{init}(\mathcal{W}, \mathcal{P}) \in \mathcal{R}^3$   ▷ Initialization
7: **for** n in N **do**
8:     $p \leftarrow \frac{\sum_{p_i, w_i \in (P, W)} K(p_i - p) w_i p_i}{\sum_{p_i, w_i \in \{P, W\}} K(p_i - p) w_i}$   ▷ $K(x) = e^{-\frac{\|x\|^2}{2\sigma^2}}$
9: **end for**
10: **Output:** $p$

---

*i.e.* $K = 5$ and the kernel width $\sigma$ as 40mm based on ablative analysis.

### 3.4. Implementation Details

The network is implemented with Tensorflow[1] and optimized using the Adam [15] with the initial learning rate set to 0.001 and the exponential decay rate of the momentum $\beta_1 = 0.5$. Following [11, 24], we randomly rotate the input depth map and change the aspect ratio for data augmentation. The batch size is set as 40 and we use batch re-normalization to accelerate training, which works better on small training mini-batches compared to batch normalization [14]. During testing, we use two network stacks and have an average run time of 36m̈s per image (27.8ḞPS) on a single NVIDIA Titan X GPU card.

## 4. Experiments

We conduct experiments on 3 publicly available datasets, *i.e.* NYU[43], MSRA[37] and ICVL[39]. We choose the NYU dataset to conduct ablation experiments and compare against the baseline methods since it has a wider coverage of hand poses as opposed to the other two.

We quantitatively evaluate our method with two metrics: mean joint error (in mm) averaged over all joints and all frames, and percentage of frames in which all joints are below a certain threshold [41]. Qualitative results of the estimated hand poses are shown in Fig. 5 and Fig. 8.

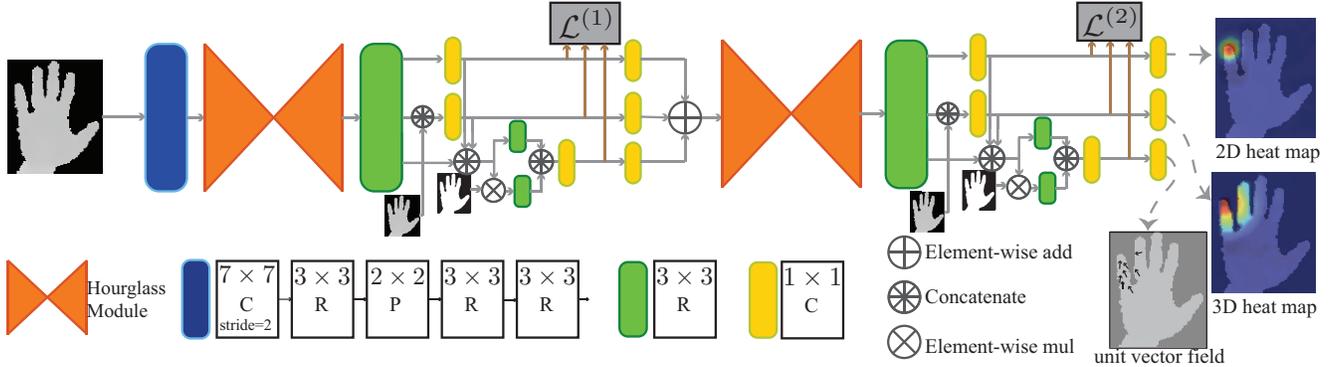

Figure 1. **Network architecture.** The abbreviations C, P, R stands for convolution layer, pooling and residual module respectively. We choose 128*128 as size of input depth map and 32*32 as the input and output resolution of hourglass module[23] with 128 feature channels in each layer. In this paper, we use 2 stacks due to real-time performance constraint. The network estimate 2D,3D heat maps and unit vector field for each joint, we only show the pinky tip point here. Figure is best viewed in colour.

### 4.1. Baseline methods

In this section we analyze whether regression of 2D joint detections helps 3D regression and how different strategies to fuse 2D joint detections and 3D regressions impact the final pose accuracy. In addition, we also show the influence of choosing different 3D offset parameterizations.

**Does 2D joint detection help with 3D regression?** First, we would like to find out if 2D joint detection actually is helpful 3D holistic regression. To that end, we design two baseline methods: directly regressing 3D joint coordinates versus coupling 2D joint detection and 3D regression in a multi-task setup. Specifically, for **baseline 1 (coordinate regression)**, the regression network follows the architecture from Fig. 2(a) which takes a depth map as input and directly outputs 3D joint coordinates. For **baseline 2 (detection+coordinate regression)**, we adopt a similar regression network architecture (see Fig. 2(b)) but add an hourglass module[23]. We feed the depth map, the feature map from the hourglass module, and the 2D joint detection heat map all concatenated together as input into the brown module in Fig. 2,and train for regression. Furthermore, to ensure a fair comparison to our proposed method, we also stack two of such networks together for baseline 2.

As is shown in Fig. 3, there is only a minor improvement of 0.16mm in terms of the average joint error from direct *coordinate regression* to *detection+coordinate regression*. Furthermore, both methods perform similarly when the error threshold is larger than 25mm. We conclude that while 2D detection may help in learning a better feature map, coupling 2D detection together with 3D regression does not solve the inherent problems of 3D regression, *e.g.*, translation variance and inability to generalize through combining local evidence.

**Impact of fusion strategies** To further explore better strategies for fusion of 2D detection and 3D regression, we design an alternative method using the identical network architecture as *detection+coordinate regression*(see Fig. 2 (b)) except for the output layer. Instead of regressing $(x, y, z)$ as per baseline 2, we regress only the $z$ coordinate, and refer to this as **baseline 3 (detection+depth regression)**. This output in the $z$ axis is then combined with the 2D detection results which are used directly as the coordinates for $x, y$ plane.

Surprisingly, *detection+depth regression* outperforms *detection+coordinate regression* both in terms of the average joint error and the percentage of frames below the error threshold from 20 to 50 mm (see Fig. 3). This suggests that 2D detection provides a more accurate estimate than coordinate regression. We conclude that it should be beneficial to explicitly enforce some form of consensus between the 3D estimates and the 2D detections. While the accuracy of this baseline is still lower than our proposed approach by a large margin (see Fig. 3), it shows that treating depth maps as 2D images and using CNNs for holistic depth regression is not enough to resolve the depth ambiguity in 3D hand pose estimation.

Besides fusing the 2D detection heat maps as input for coordinate regression, a second line of work [43, 10, 51] conducts model-based tracking based on inverse kinematics to recover the 3D pose. We compare against previous state-of-the-art methods [43, 10, 51] based on such a strategy and out-perform all of them (see Section 4.3). This validates the effectiveness of our proposed method in handling depth ambiguities arising from the severe self occlusions in the hand.

**Impact of offset re-parameterization** We implement a network which directly regresses the 3D offset without re-parameterization into the 3D heatmap and directional unit-vector as **baseline 4 (mask loss)**. As is shown in Fig. 2 (c), the offset regression architecture follows exactly the same structure as the offset unit direction regression in Fig. 1.

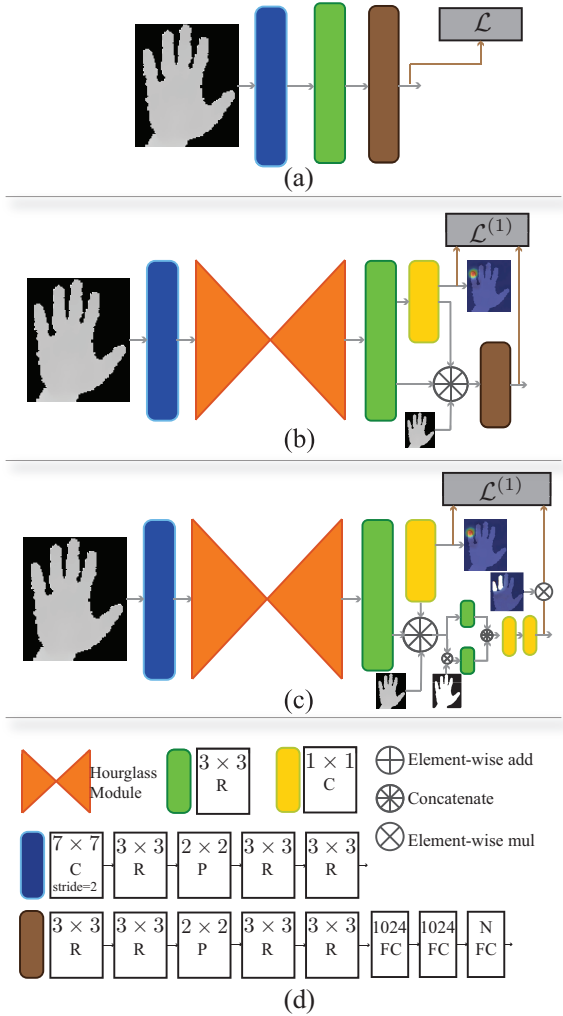

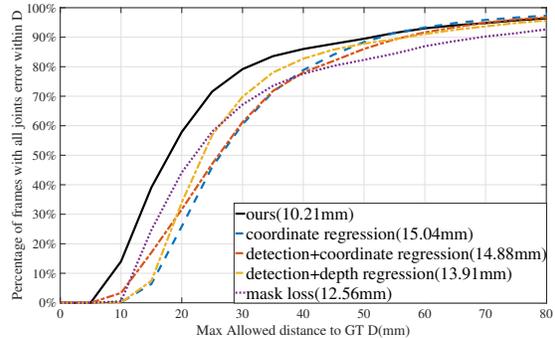

Figure 2. **Baseline network architectures.** (a) Direct 3D coordinate regression from depth map (baseline 1); (b) Network regresses 3D joint coordinates (baseline 2) or z-axis coupled 2D joint detection (baseline 3) together with the 2D joint detection heatmaps; (c) Regressing 3D offset vector field by masking the loss with the 3D distance to joint (baseline 4); (d) Detailed architecture configurations. The abbreviations C, P, R, FC stands for convolutional layer, pooling, residual module, and fully connected layer respectively. For (b) and (c), we experiment with a stack of 2 in the same way as the proposed method for fair comparison.

We use the 2D detection scores to select candidate points as inputs to the mean shift. In this baseline, we apply a 3D distance threshold to the loss function of the offset, as was done in [28], and effectively masks the regression so that we only regress a joint's neighbour points. Ideally, this baseline should be conducted without masking, but the training failed completely, with the loss oscillating back and forth without decreasing. As is shown in Fig. 3, pixel-wise dense estimation out-performs holistic regression method and validates the benefits of regressing point-wise 3D offsets.

Figure 3. **Comparison with baselines.** We compare our approach to four baseline methods (Sec. 4.1) on the NYU dataset[43]. Number in the parenthesis of the legend indicates the average 3D error of the corresponding method.

Given the insights drawn from these baseline experiments, we attribute the high accuracy achieved by our method to the reparameterization and imposing the loss to all points for vector field regression. Decomposing the 3D offsets into the joint 3D heat map and offset direction and regressing the two in a cascaded way is easier to learn than directly regressing the offsets. Secondly, setting the offset vector to zero for outlier points instead of excluding them from the loss makes the estimation more robust to errors in regression during testing.

### 4.2. Exploration studies

We first experiment on the number of stacked networks and the hyperparameters of mean-shift, *i.e.*, the number of selected candidate points as input to the mean shift and the kernel width. As indicated in Table 1, we find that the proposed method is quite robust to the mean shift hyperparameters. On the other hand, the number of network stacks is critical to the estimation accuracy. We test only up to 2 stacks to maintain real-time performance; however, as already shown in [23, 47], adding more stacks could improve the accuracy.

In addition, as shown in the last two rows of Table 1, the un-weighted mean shift approximation has a similar accuracy as the weighted version, with only 0.09mm difference with respect to the mean joint error. As such, we choose 2 stacks and 5 candidate points as input to mean shift, kernel width $\sigma = 40$mm and weighted mean shift as described in Alg. 1 in the following experiments.

### 4.3. Comparison to state-of-the-art

**NYU Dataset** The NYU hand dataset [43] contains over 72K training and 8K testing frames. Its wide coverage of hand poses and noisy input depths make this dataset quite challenging. Since the hand region is not cropped out, we use an hourglass joint detector [23] with one stack to locate the joints and take the median of estimated x and y coordinates over all joints respectively as the center point for cropping out the hand region. We only use *view 1* for both training and testing and evaluate on a subset of 14 joints as in [43] to make a fair comparison.

| Network parameters | | |
|---|---|---|
| # Stacks $T$ | 1 | 2 |
| | 11.20 | 10.21 |

| Mean-shift parameters | | | | |
|---|---|---|---|---|
| # Candidates $K$ | 1 | 5 | 10 | 30 |
| | 10.6 | 10.21 | 10.21 | 10.96 |
| Kernel width $\sigma$ | 10mm | 40mm | 80mm | 100mm |
| | 10.35 | 10.21 | 10.21 | 10.21 |
| Weights | weighted | | unweighted | |
| | 10.21 | | 10.26 | |

Table 1. **Impact of hyperparameters.** We report the mean 3D error (in mm) averaged over all joints and all frames on NYU dataset[43]. We choose 2 stacks and 5 nearest points as input to mean shift, kernel width $\sigma = 40mm$ and weighted mean shift as described in Alg. 1 as the default parameters.

| Method | Average 3D error |
|---|---|
| Xu et al. [49] (Lie-X) | 14.5mm |
| Wan et al. [45] (Crossing Nets) | 15.5mm |
| Oberweger et al.[24] (DeepPrior++) | 12.3mm |
| Guo et al.[12] (REN) | 12.7mm |
| Chen et al.[5] (Pose Guided) | 11.8mm |
| Ours | **10.2mm** |

Table 2. **Comparison with state-of-the-art on NYU.** We report average 3D error on the NYU[43] dataset.

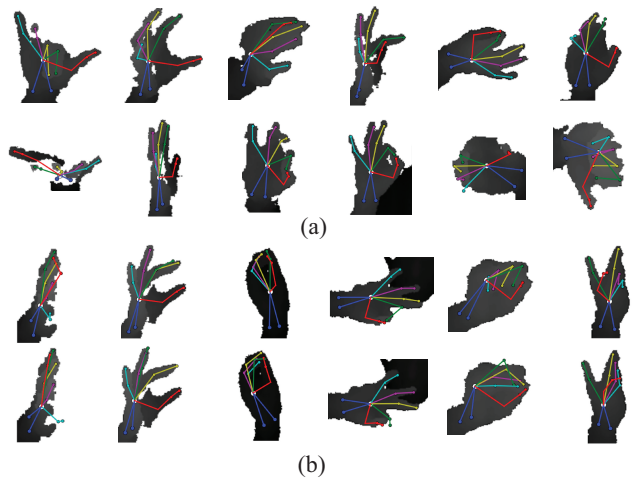

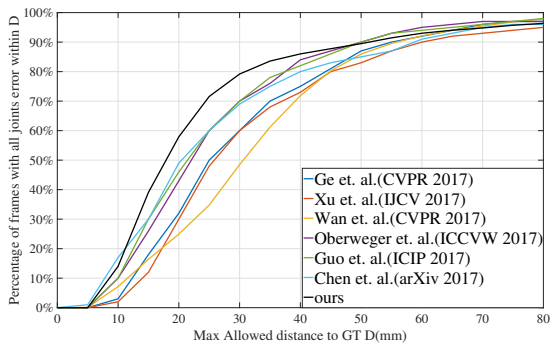

Figure 4. **Comparison with state-of-the-art on NYU [43].** We plot the percentage of frames in which all joints are below a threshold.

Figure 5. **Qualitative results.** Hand pose estimation results on NYU dataset[43]. (a) Successful samples with largest joint error below 20mm; (b) Failed samples (top row) and the corresponding ground-truth(bottom row).

We compare our method to the most recently proposed methods [49, 45, 24, 11, 12, 5]. All are 3D regression-based methods with sophisticated network architectures and surpass earlier works [43, 51, 40, 7, 25, 26, 54] by a large margin. As is shown in Fig. 4 and Tab. 2, our method outperforms all these state-of-the-art methods with a large margin for both metrics. Specifically, according to Fig. 4, our method significantly increases the percentage of successfully estimated frames by 8% (from 50% to 58%) on the error threshold of 20mm and by 9.2% (from 70% to 79.2%) on 30mm when compared to most accurate methods published to date ( [5] and [24, 12, 5] respectively). We also show qualitative results on Fig. 5. The main reasons for the failure cases are severe self occlusions and noise in the depth map.

**MSRA Dataset** The MSRA hand dataset[37] contains 76.5K images from 9 subjects with 17 hand gestures. Following the protocol of [37], we use a leave-one-subject-out training / testing split and average the results over the 9 subjects. We compare our methods with state-of-art methods[11, 37, 46, 45, 24, 12, 5]. Specifically, [46, 37] are based on the hierarchical regression forest. Similar to our approach, [46, 37] regress 3D offsets and aggregate local estimations with the mean-shift algorithm. [11, 45, 24, 12, 5] are CNN based 3D holistic regression methods and outperforms other existing methods[10, 18].

Again, our method outperforms all state-of-the-art by a large margin both in terms of percentage of successful frames (see Fig. 6) and average joint error (see Tab. 3). As is shown in Fig. 6, over 81% and 91% of frames have joint errors below 20mm and 30mm. This is a huge improvement over the most accurate existing results from [5], which has only 60% and 81% respectively. The qualitative results is shown in Fig. 8(b).

**ICVL Dataset** The ICVL hand dataset[39] has 22K frames for training and 1.5k for testing. An additional 160k augmented frames with in-plane rotations are provided by the dataset but we do not use them as we perform data augmentation on the fly during training as described in Sec. 3.4. The variance in pose is much smaller in ICVL compared to the NYU and MSRA datasets. We compare our method against [37, 46, 40, 45, 24, 12, 5]. [37, 46, 40] are based on hierarchical regression forest and others [45, 24, 12, 5] on 3D holistic regression.

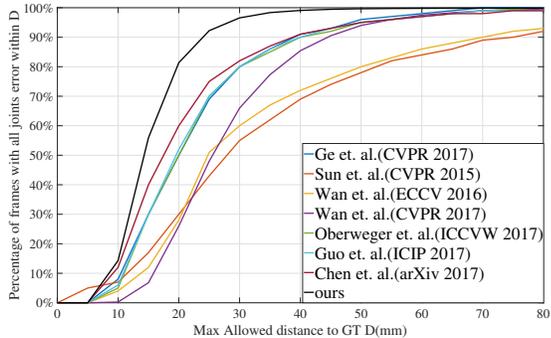

Figure 6. Comparison with state-of-the-art on MSRA[37]. We plot the percentage of frames in which all joints are below a threshold.

| Method | Average 3D error |
| --- | --- |
| Ge et al. [11] (3D CNN) | 9.5mm |
| Wan et al.[45] (Crossing Nets) | 12.2mm |
| Oberweger et al.[24] (DeepPrior++) | 9.5mm |
| Guo et al.[12] (REN) | 9.8mm |
| Chen et al. [5] (Pose Guided) | 8.6mm |
| Ours | **7.2mm** |

Table 3. **Comparison with state-of-art on MSRA [37].** We plot the percentage of frames in which all joints are below a threshold.

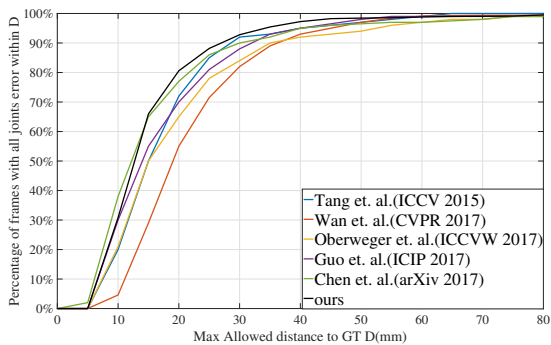

Figure 7. **Comparison with state-of-the-art on ICVL [39].** We plot the percentage of frames in which all joints are below a threshold.

As is shown in Fig. 7, our method achieves similar accuracy as [5] and outperforms the rest. Our method has an average 3D error on par with [5] and better than the others. We consider the differences between our method and [5] as being less significant given the result is nearly saturated. The qualitative results can be seen in Fig. 8.

| Method | Average 3D error |
| --- | --- |
| Wan et al.[45] (Crossing Nets) | 10.2mm |
| Wan et al. [46] (Surface Normal) | 8.2mm |
| Sun et al. [37] (Cascaded Regression) | 9.9mm |
| Oberweger et al.[24] (DeepPrior++) | 8.1mm |
| Guo et al.[12] (REN) | 7.5mm |
| Chen et al. [5] (Pose Guided) | **6.8mm** |
| Ours | 7.3mm |

Table 4. Comparison with state-of-art on ICVL[39] dataset.

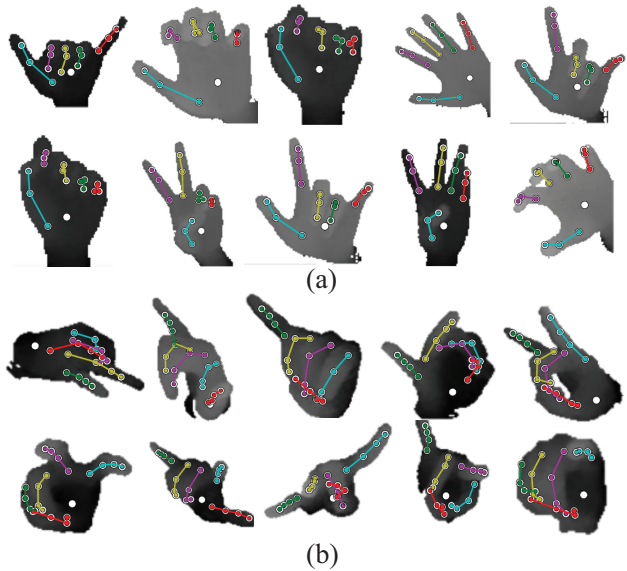

(a)

(b)

Figure 8. **Qualitative results.** Hand pose estimation results from (a) ICVL[39], (b) MSRA[37].

## 5. Conclusion and discussion

We propose a highly accurate method for 3D hand pose estimation from single depth map inputs. Given a depth camera frame, we decompose 3D pose parameters into a set of 2D/3D joint heat maps and 3D unit vector fields of offset directions. This reparameterization allows us to consider both the 2D and 3D properties of the depth map and makes it easy to leverage fully convolutional networks. We aggregate local estimations by a non-parametric mean shift variant, which explicitly enforces the estimated 3D joint coordinates to be in accordance with the 2D and 3D local estimations. Our method provides a better fusion scheme between 2D detection and 3D regression than previous state-of-the-art and the various baselines. As future work, we plan to further extend our method for 3D pose estimation from RGB inputs as well as for hands grasping objects.

## References

[1] M. Abadi, A. Agarwal, P. Barham, E. Brevdo, Z. Chen, C. Citro, G. S. Corrado, A. Davis, J. Dean, M. Devin, et al. Tensorflow: Large-scale machine learning on heterogeneous

distributed systems. *arXiv preprint arXiv:1603.04467*, 2016. 4

[2] F. Bogo, A. Kanazawa, C. Lassner, P. Gehler, J. Romero, and M. J. Black. Keep it SMPL: Automatic estimation of 3D human pose and shape from a single image. In *ECCV*, 2016. 2

[3] Z. Cao, T. Simon, S.-E. Wei, and Y. Sheikh. Realtime multi-person 2D pose estimation using part affinity fields. In *CVPR*, 2017. 3

[4] C.-H. Chen and D. Ramanan. 3D Human Pose Estimation = 2D Pose Estimation + Matching. In *CVPR*, 2017. 2

[5] X. Chen, G. Wang, H. Guo, and C. Zhang. Pose guided structured region ensemble network for cascaded hand pose estimation. *arXiv preprint arXiv:1708.03416*, 2017. 1, 3, 7, 8

[6] C. Choi, S. Kim, and K. Ramani. Learning hand articulations by hallucinating heat distribution. In *ICCV*, 2017. 3

[7] C. Choi, A. Sinha, J. H. Choi, S. Jang, and K. Ramani. A collaborative filtering approach to Real-Time hand pose estimation: Supplementary material. In *ICCV*, 2015. 7

[8] J. Dai, K. He, and J. Sun. Instance-aware semantic segmentation via multi-task network cascades. In *CVPR*, 2016. 4

[9] E. Dibra, T. Wolf, C. Oztireli, and M. Gross. How to Refine 3D Hand Pose Estimation from Unlabelled Depth Data? In *3DV*, 2017. 3

[10] L. Ge, H. Liang, J. Yuan, and D. Thalmann. Robust 3D hand pose estimation in single depth images: from single-view cnn to multi-view cnns. In *CVPR*, 2016. 1, 2, 3, 5, 7

[11] L. Ge, H. Liang, J. Yuan, and D. Thalmann. 3D convolutional neural networks for efficient and robust hand pose estimation from single depth images. In *CVPR*, 2017. 2, 3, 4, 7, 8

[12] H. Guo, G. Wang, X. Chen, C. Zhang, F. Qiao, and H. Yang. Region ensemble network: Improving convolutional network for hand pose estimation. In *ICIP*, 2017. 3, 7, 8

[13] K. He, G. Gkioxari, P. Dollár, and R. Girshick. Mask R-CNN. In *ICCV*, 2017. 4

[14] S. Ioffe. Batch renormalization: Towards reducing minibatch dependence in batch-normalized models. *arXiv preprint arXiv:1702.03275*, 2017. 4

[15] D. Kingma and J. Ba. Adam: A method for stochastic optimization. In *ICLR*, 2015. 4

[16] I. Lifshitz, E. Fetaya, and S. Ullman. Human pose estimation using deep consensus voting. In *ECCV*, 2016. 3

[17] M. Lin, L. Lin, X. Liang, K. Wang, and H. Cheng. Recurrent 3D pose sequence machines. In *CVPR*, 2017. 1, 2

[18] M. Madadi, S. Escalera, A. Carruesco, C. Andujar, X. Baró, and J. Gonzàlez. Occlusion aware hand pose recovery from sequences of depth images. In *FG*, 2017. 7

[19] J. Martinez, R. Hossain, J. Romero, and J. J. Little. A simple yet effective baseline for 3D human pose estimation. In *ICCV*, 2017. 2

[20] F. Milletari, S.-A. Ahmadi, C. Kroll, A. Plate, V. Rozanski, J. Maiostre, J. Levin, O. Dietrich, B. Ertl-Wagner, K. Bötzel, et al. Hough-cnn: Deep learning for segmentation of deep brain regions in mri and ultrasound. *CVIU*, 2017. 3

[21] G. Moon, J. Y. Chang, Y. Suh, and K. M. Lee. Holistic planimetric prediction to local volumetric prediction for 3d human pose estimation. *arXiv preprint arXiv:1706.04758*, 2017. 2

[22] F. Moreno-Noguer. 3D human pose estimation from a single image via distance matrix regression. In *CVPR*, 2017. 2, 3

[23] A. Newell, K. Yang, and J. Deng. Stacked hourglass networks for human pose estimation. In *ECCV*, 2016. 1, 2, 3, 4, 5, 6

[24] M. Oberweger and V. Lepetit. Deepprior++: Improving fast and accurate 3D hand pose estimation. In *ICCVW*, 2017. 3, 4, 7, 8

[25] M. Oberweger, P. Wohlhart, and V. Lepetit. Hands deep in deep learning for hand pose estimation. *arXiv preprint arXiv:1502.06807*, 2015. 1, 3, 7

[26] M. Oberweger, P. Wohlhart, and V. Lepetit. Training a feedback loop for hand pose estimation. In *ICCV*, 2015. 3, 7

[27] I. Oikonomidis, N. Kyriazis, and A. A. Argyros. Efficient model-based 3D tracking of hand articulations using kinect. In *BMVC*, 2011. 1

[28] G. Papandreou, T. Zhu, N. Kanazawa, A. Toshev, J. Tompson, C. Bregler, and K. Murphy. Towards accurate multi-person pose estimation in the wild. In *CVPR*, 2017. 3, 4, 6

[29] G. Pavlakos, X. Zhou, K. G. Derpanis, and K. Daniilidis. Coarse-to-fine volumetric prediction for single-image 3D human pose. In *CVPR*, 2017. 2

[30] A.-I. Popa, M. Zanfir, and C. Sminchisescu. Deep multitask architecture for integrated 2D and 3D human sensing. In *CVPR*, 2017. 1, 2

[31] C. Qian, Q. Chen, S. Xiao, W. Yichen, T. Xiaoou, and S. Jian. Realtime and robust hand tracking from depth. In *CVPR*, 2014. 1

[32] G. Rogez, P. Weinzaepfel, and C. Schmid. LCR-Net: Localization-classification-regression for human pose. In *CVPR*, 2017. 1, 2

[33] J. Shotton, R. Girshick, A. Fitzgibbon, T. Sharp, M. Cook, M. Finocchio, R. Moore, P. Kohli, A. Criminisi, A. Kipman, and A. Blake. Efficient human pose estimation from single depth images. *TPAMI*, 2013. 1, 2, 3

[34] H. Sidenbladh, M. J. Black, and D. J. Fleet. Stochastic tracking of 3D human figures using 2D image motion. In *ECCV*, 2000. 2

[35] L. Sigal, A. O. Balan, and M. J. Black. Humaneva: Synchronized video and motion capture dataset and baseline algorithm for evaluation of articulated human motion. *IJCV*, 87(1):4–27, 2010. 2

[36] X. Sun, J. Shang, S. Liang, and Y. Wei. Compositional human pose regression. In *ICCV*, 2017. 3

[37] X. Sun, Y. Wei, S. Liang, X. Tang, and J. Sun. Cascaded hand pose regression. In *CVPR*, 2015. 2, 3, 4, 7, 8

[38] J. S. Supancic, G. Rogez, Y. Yang, J. Shotton, and D. Ramanan. Depth-Based hand pose estimation: Data, methods, and challenges. In *ICCV*, 2015. 3

[39] D. Tang, H. J. Chang, A. Tejani, and T.-K. Kim. Latent regression forest: Structured estimation of 3D articulated hand posture. In *CVPR*, 2014. 2, 4, 7, 8

[40] D. Tang, J. Taylor, P. Kohli, C. Keskin, T.-K. Kim, and J. Shotton. Opening the black box: Hierarchical sampling optimization for estimating human hand pose. In *ICCV*, 2015. 2, 3, 7, 8

[41] J. Taylor, J. Shotton, T. Sharp, and A. Fitzgibbon. The vitruvian manifold: Inferring dense correspondences for one-shot human pose estimation. In *CVPR*, 2012. 4

[42] D. Tome, C. Russell, and L. Agapito. Lifting from the deep: Convolutional 3D pose estimation from a single image. In *CVPR*, 2017. 1, 2

[43] J. Tompson, M. Stein, Y. Lecun, and K. Perlin. Real-time continuous pose recovery of human hands using convolutional networks. *ACM Transactions on Graphics (TOG)*, 2014. 1, 2, 3, 4, 5, 6, 7

[44] A. Toshev and C. Szegedy. DeepPose: Human pose estimation via deep neural networks. In *CVPR*, 2014. 1

[45] C. Wan, T. Probst, L. Van Gool, and A. Yao. Crossing nets: Combining GANs and VAEs with a shared latent space for hand pose estimation. In *CVPR*, 2017. 3, 7, 8

[46] C. Wan, A. Yao, and L. Van Gool. Hand pose estimation from local surface normals. In *ECCV*, 2016. 2, 3, 7, 8

[47] S.-E. Wei, V. Ramakrishna, T. Kanade, and Y. Sheikh. Convolutional pose machines. In *CVPR*, 2016. 1, 2, 4, 6

[48] Y. Xie, X. Kong, F. Xing, F. Liu, H. Su, and L. Yang. Deep voting: A robust approach toward nucleus localization in microscopy images. In *International Conference on Medical Image Computing and Computer-Assisted Intervention*, 2015. 3

[49] C. Xu, L. N. Govindarajan, Y. Zhang, and L. Cheng. Lie-X: Depth image based articulated object pose estimation, tracking, and action recognition on lie groups. *IJCV*, 2017. 3, 7

[50] A. Yao, J. Gall, L. V. Gool, and R. Urtasun. Learning probabilistic non-linear latent variable models for tracking complex activities. In *NIPS*, 2011. 2

[51] Q. Ye, S. Yuan, and T.-K. Kim. Spatial attention deep net with partial PSO for hierarchical hybrid hand pose estimation. In *ECCV*, 2016. 1, 2, 3, 5, 7

[52] S. Yuan, Q. Ye, B. Stenger, S. Jain, and T.-K. Kim. Bighand2.2m benchmark: Hand pose dataset and state of the art analysis. In *CVPR*, 2017. 1

[53] X. Zhou, Q. Huang, X. Sun, X. Xue, and Y. Wei. Towards 3D human pose estimation in the wild: A weakly-supervised approach. In *ICCV*, 2017. 2

[54] X. Zhou, Q. Wan, W. Zhang, X. Xue, and Y. Wei. Model-based deep hand pose estimation. In *IJCAI*, 2015. 3, 7